\documentclass[12pt, draftclsnofoot, onecolumn]{IEEEtran}
\usepackage[T1]{fontenc}

\usepackage{xcolor}

\usepackage{cite}

\ifCLASSINFOpdf
   \usepackage[pdftex]{graphicx}
  \usepackage{subcaption}
\else
\fi
\usepackage{amsmath}
\usepackage[cmintegrals]{newtxmath}
\usepackage[ruled]{algorithm2e}

\usepackage{tabularx}

\usepackage{caption}
\begin{document}
%
\title{Adaptive Early Exiting for Collaborative\\Inference over Noisy Wireless Channels}
%
%
%

\author{Mikolaj Jankowski, Deniz G{\"u}nd{\"u}z and Krystian Mikolajczyk\\
Imperial College London}

\maketitle

\begin{abstract}
Collaborative inference systems are one of the emerging solutions for deploying deep neural networks (DNNs) at the wireless network edge. Their main idea is to divide a DNN into two parts, where the first is shallow enough to be reliably executed at edge devices of limited computational power, while the second part is executed at an edge server with higher computational capabilities. The main advantage of such systems is that the input of the DNN gets compressed as the subsequent layers of the shallow part extract only the information necessary for the task. As a result, significant communication savings can be achieved compared to transmitting raw input samples. In this work, we study early exiting in the context of collaborative inference, which allows obtaining inference results at the edge device for certain samples, without the need to transmit the partially processed data to the edge server at all, leading to further communication savings. The central part of our system is the transmission-decision (TD) mechanism, which, given the information from the early exit, and the wireless channel conditions, decides whether to keep the early exit prediction or transmit the data to the edge server for further processing. In this paper, we evaluate various TD mechanisms and show experimentally, that for an image classification task over the wireless edge, proper utilization of early exits can provide both performance gains and significant communication savings.
\end{abstract}

\begin{IEEEkeywords}
Collaborative learning, edge machine learning, joint source-channel coding, early exit.
\end{IEEEkeywords}

%
\IEEEpeerreviewmaketitle

\section{Introduction}
The rise of the fifth-generation (5G) standard for broadband cellular networks drives the transformation of how we understand connectivity and apply it in our everyday lives. The wide availability of reliable wireless connectivity allows many new technologies to be delivered closer to end users. This includes video streaming, augmented reality (AR), various Internet of Things (IoT) applications, among others.

One of the most emerging application areas, where the users can significantly benefit from the improved connectivity is edge machine learning (ML). The majority of edge devices are unable to run complex ML algorithms, such as deep neural networks (DNNs), due to either computational power or memory limitations. Currently, edge devices offload their data to a powerful \textit{edge server} through a wireless link. The edge server can then run the desired ML algorithm on the delivered data, and return the inference result to the device.

\begin{figure*}[t]
    \centering
    \includegraphics[width=0.85\textwidth]{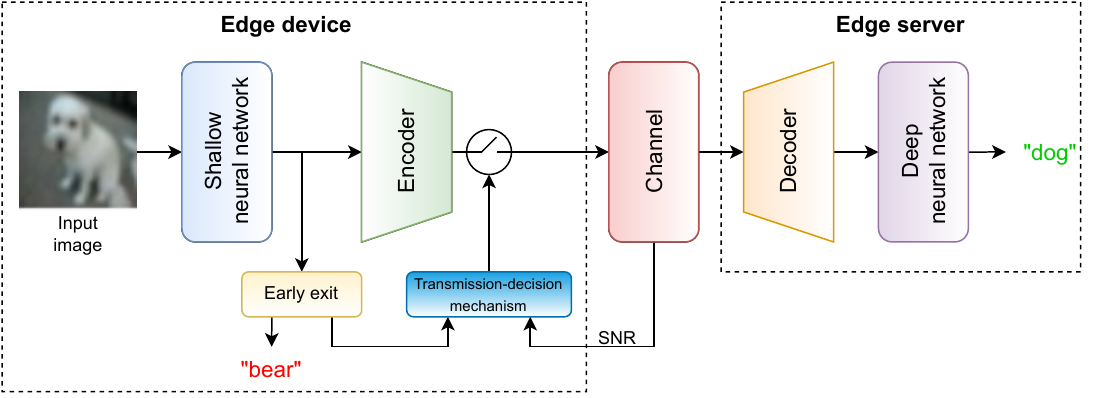}
    \caption{System model. The edge device performs a forward pass of a shallow NN. Early exit attempts to make a prediction based on the output of NN, and passes its result to a TD mechanism, which combines this result with the current channel state to decide whether to transmit the output of the shallow NN to the edge server, where additional processing is performed by a DNN to reach a more refined prediction.}
    \label{fig:system_model}
\end{figure*}

Collaborative inference is a recent extension to the edge ML paradigm \cite{collaborative_inference_survey, gunduz_magazine}. In collaborative inference, a complex ML algorithm is split into two parts between the edge device and the edge server. The edge device utilizes its limited computational power to run the initial part of the algorithm, i.e., the first layers of the DNN, and offloads the intermediate result to the edge server, which runs the remainder of the algorithm, and provides the final result to the edge device. Deep neural networks are particularly suitable for collaborative inference thanks to their sequential multi-layered structure, which provides many potential splitting points. An important benefit of executing an initial part of the DNN on the edge device is that DNNs trained for a specific task learn to only preserve the information essential for that task. Therefore, each layer of a DNN reduces the amount of information contained in the intermediate results, denoted as intermediate feature maps. As a result, executing even a few layers of a DNN on the edge device can significantly reduce the amount of information to be transmitted to the edge server. Such collaborative inference systems for the wireless edge have been extensively studied in \cite{jankowski_spawc, bottlenet, bn_plus_plus, distributed_inference, early_exit_computations}.

In the case of constantly changing wireless channel conditions, it may not always be the best choice to offload the intermediate feature maps to the edge server for inference. When the channel conditions are poor, the delivered features may become too distorted to allow for successful inference at the edge server. Moreover, certain input samples may require less processing than others, making edge processing unnecessary. In this work, we study the application of early exits in the context of wireless collaborative inference at the edge. The idea of early exits, initially proposed in \cite{googlenet}, is to attach additional outputs to a DNN, which allow obtaining earlier predictions without the need to execute all the layers of the DNN. Originally, early exits were proposed as a solution to vanishing gradient problems. Since then, they have been extensively studied for efficient ML as a method for avoiding excessive computations \cite{branchynet_early_exit, adaptive_early_exit, early_exit_why, marcello_efficient_ml}. We note that, in the collaborative inference problem, early exits can allow the edge device to obtain an inference result locally, without the need to transmit the intermediate features to the edge server, which has been noticed in \cite{early_exit_edge_survey}, while \cite{split_early_exit} introduced a scheme for dynamic early exiting based on classification confidence thresholds. The benefits of such an approach are communication savings, reduced computations at the edge server, and better inference performance in the presence of a low channel quality between the device and the edge server.

We consider a collaborative inference system (see Fig. \ref{fig:system_model}), where an edge device can decide whether it should rely on its own inference following an early exit, or transmit the obtained features to the edge server. Optimizing such a decision policy is challenging as it has to rely solely on the information available at the edge device, i.e., the available features and channel state information (CSI), and with the lowest possible computational cost. We propose an automated mechanism for making this decision, based on a lightweight neural network, which analyzes the data available from the early exit combined with the CSI. To the best of our knowledge, this is the first work to analyze early exits in the context of wireless edge collaborative inference systems, and one of the first (similar NNs for accepting early exit output were proposed in \cite{learnable_early_exit_eenet, learnable_early_exit_epnet} in the context of efficient DNN inference) to study a DL-based decision mechanism for accepting early exit inference results. We provide a systematic comparison of such mechanisms.

\section{Methods}
\subsection{System model}

The collaborative inference system studied in this work consists of a DNN for image classification, which is split into two parts (see Fig. \ref{fig:system_model}). The first part is deployed at the edge device, and the second at the edge server. The intermediate features produced by the first part of the DNN are compressed by a joint source-channel coding (JSCC) encoder neural network, which maps the intermediate features directly into channel input symbols, and follows the architecture proposed in \cite{jankowski_spawc}. These symbols are then transmitted through a wireless channel to the edge server, where they are reconstructed by a JSCC decoder neural network and further processed by the remaining part of the image classification DNN. Early exit layers $F_{e}$ are added at the splitting point of the DNN, in order to obtain early predictions for the underlying task. We note, that obtaining correct prediction through early exit allows the edge device to avoid transmitting intermediate features to the receiver, thus saving communication resources.
We introduce a transmission-decision (TD) mechanism for obtaining a decision on whether the edge device should stop at the early exit prediction, or transmit the intermediate features to the edge server for further processing. The goal of a TD mechanism is to maximize the classification accuracy as well as \textit{communication savings}, defined as a fraction of images classified at the edge device by the early exit, without transmitting the intermediate feature maps to the edge server for further processing. For example, transmission savings of 0.4 indicate that $40\%$ of images were classified locally, while the remaining $60\%$ required transmitting features to the edge server. Alongside several simple TD mechanisms, we design a transmission-decision NN (TD NN) to learn the correct decision policy based on the output of the early exit and the channel state information. We note that this problem is multi-faceted, as one has to consider not only the expected performances of early and final exits, but also the cost of transmitting the intermediate features, as well as their quality at the receiver, given the instantaneous quality of the wireless channel between the edge device and the edge server.

\subsection{Channel model}
We consider additive white Gaussian noise (AWGN) channel between the edge device and the edge server, modeled as $\mathbf{y}=\mathbf{x}+\mathbf{z}$, where $\mathbf{x} \in \mathbb{R}^B$ is the channel input vector and $\mathbf{z}$ is the additive noise vector with each component independently drawn from the zero-mean Gaussian distribution with variance $\sigma^2$. Here, $B$ denotes the available channel bandwidth to transmit the intermediate features to the edge server. An average power constraint is imposed such that $P=\frac{1}{B}\sum_{i=1}^B x_i^2 \leq 1$; thus, the channel signal-to-noise ratio ($\mathrm{SNR}$) can be calculated as $\mathrm{SNR}=\frac{P}{\sigma^2}$.

\subsection{Training strategy}
\label{subs:training_strategy}
The training strategy applied to train each part of the proposed system is as follows. In the first step, we initialize the DNN and the early exit layers. The DNN is trained with the sum of the cross-entropy losses between the ground truth classification labels and the predictions of both early exit $F_{e}$ at the edge and final exit $F_{f}$ at the server. Please note that splitting of the DNN is not yet performed at this step, as we pre-train the entire DNN for the image classification task first. In the second step, we split the DNN, and introduce a JSCC autoencoder to communicate over the wireless channel model at the splitting point. We then train the resulting architecture by first freezing the image classification DNN and training only the JSCC autoencoder. Once the JSCC autoencoder converges, we unfreeze the DNN and train the whole architecture end-to-end with the same loss function as in the previous step. Finally, we introduce the TD NN and train it until convergence with each of the strategies detailed in Section \ref{subs:decision_network}. Alongside the TD NN, we study alternative TD mechanisms, which do not require training.

\subsection{Transmission-decision neural network (TD NN)}
\label{subs:decision_network}
\begin{figure}[t]
    \centering
\includegraphics[width=0.7\textwidth]{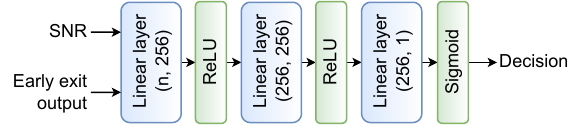}
    \caption{The architecture of the TD NN. The DNN considers the outputs of the early exit and the CSI to obtain the decision on whether to accept the early exit predictions or transmit the intermediate feature maps to the edge server for further processing. The numbers in parenthesis indicate the number of input and output features for each linear layer.}
    \label{fig:td_nn_architecture}
\end{figure}

The TD NN $D: \mathbb{R}^n\xrightarrow{}[0, 1]$, where $n$ is the number of features input to the TD NN. This number may vary based on the selection of the TD NN inputs, which may include outputs of the early exit, statistics of these outputs, or channel SNR. The choice of these inputs is analyzed in Section \ref{subs:ablation_study}. TD NN is trained to make a decision on whether the edge device should transmit the intermediate features to the edge server for further processing, or output the prediction made by the early exit. Overall, defining a proper training criterion for the TD NN is a non-trivial task. On one hand, the goal is to maximize the classification accuracy; on the other hand, the system should also minimize the communication costs, that is, should transmit only when the expected performance gain given by the final exit is sufficiently large. To account for all these criteria, we consider multiple training objectives for the TD NN and carefully study their impact on the final performance in terms of the trade-off between accuracy and communication savings. The final decision on whether to transmit the features or keep the early exit predictions should be a binary choice; therefore, during inference, we apply rounding operation to the TD NN outputs. The architecture of the TD NN is relatively simple, consisting of 3 linear layers with ReLU activations and 256-dimensional hidden features, as shown in Fig. \ref{fig:td_nn_architecture}.

\textbf{Joint early-final cross-entropy criterion.} The first training criterion we consider uses the weighted combination of the predictions made by the early exit and the final exits of the DNN, where the TD NN output is used as the weights. We define the joint early-final exit prediction as follows:
\begin{equation}
    F(x) \triangleq D(x)\cdot F_e(x) + (1 - D(x)) \cdot F_f(x),
    \label{eq:joint_prediction}
\end{equation}
where $x$ is the input sample. Then, the loss function is given by:
\begin{equation}
    L_{joint} = L_{ce}(F(x), y_{gt}) + \beta \cdot |1-D(x)|,
    \label{eq:joint}
\end{equation}
where $L_{ce}$ is the cross-entropy loss, $y_{gt}$ is the classification ground truth one-hot vector for the input sample $x$, and $\beta$ is a weighting parameter used to balance the accuracy with the communication cost. Note that, in (\ref{eq:joint}), the first term is responsible for the accuracy of the joint early-final exit prediction, while the second is a penalty term that is aimed at prioritizing early exit and avoiding transmission.

We note that the decision is a binary choice, i.e., $D(x): \mathbb{R}^n\xrightarrow{}\{0, 1\}$, while the outputs of the TD NN can be any value between 0 and 1. Therefore, during the training we calculate the decision as $D(x) = \frac{1}{1 + e^{-T\cdot \hat{D}(x)}}$, where $\hat{D}(x)$ are the raw outputs of the TD NNs, and $T$ is the temperature parameter used to push the decisions away from ambiguous values close to $0.5$.

\textbf{Binary cross-entropy criterion.} The second criterion used for training TD NN is based on the assumption that the optimal decision is to transmit intermediate features only when the early exit prediction is likely to be incorrect, while the final exit prediction is likely to be correct. Based on this assumption, we generate a set of ground truth decisions $d_{gt}$, such that $d_{gt}=1$ if and only if the early exit is incorrect and the final exit is correct, and $d_{gt}=0$ otherwise. We then train the TD NN with the following loss function:
\begin{equation}
    L_{gt} = L_{bce}(D(x), d_{gt}) + \beta \cdot |1-D(x)|,
    \label{eq:gt_decision}
\end{equation}
where $L_{bce}$ is the binary cross-entropy function, and the second term is for penalizing excessive transmissions as before.

\textbf{Mixed criterion.} The final criterion we use to train the TD NN is a mixed approach. It includes a combination of the two aforementioned criteria to obtain the following loss function:
\begin{equation}
    L_{mixed} = L_{ce}(F(x), y_{gt}) + \alpha L_{bce}(D(x), d_{gt}) + \beta \cdot |1-D(x)|.
    \label{eq:mixed}
\end{equation}
\section{Results}

\subsection{Experimental setup}
\label{subs:experimental_setup}
We evaluate the proposed approach on the CIFAR100 dataset \cite{cifar}, which consists of 60000 images of 100 distinct classes. Each class is represented with 500 samples in the training set and 100 samples in the test set. We employ the VGG16 architecture \cite{vgg}. To avoid excessive computations on the edge device, the early exit consists of an average pooling operation, a linear layer with ReLU activation, followed by another linear layer with softmax activation, which outputs 100 class predictions.

The training of the DNN follows the strategy presented in Section \ref{subs:training_strategy}. We utilize the stochastic gradient descent (SGD) optimizer with a batch size of 128, and run 90, 30, and 30 epochs for each training step, respectively. In the first training step, we set the learning rate to $0.1$, and decay it by a factor of 10 every 30 epochs. In the second and third training steps, we also start from a learning rate of $0.1$, but the decay occurs every 10 epochs. At each training iteration, starting from the second training step, we set the channel bandwidth $B$ to 64 channel uses and determine the channel SNR according to the ``sandwich rule'' \cite{slimmable_nets}, that is, we set $\mathrm{SNR} = -10\,\mathrm{dB}$ in the first iteration, $\mathrm{SNR} = 10\,\mathrm{dB}$ in the second iteration, and, in the third iteration, $\mathrm{SNR} = \mathcal{U}(-10\,\mathrm{dB}, 10\,\mathrm{dB})$, where $\mathcal{U}(\cdot, \cdot)$ represents the uniform distribution. This cycle is then repeated throughout the whole training. This ensures that the system learns to handle diverse channel conditions by training a single set of DNNs exposed to a large SNR variability throughout training. An alternative approach would be to train a separate set of DNNs for each channel SNR, which imposes a significant memory overhead and is not considered in this paper.
For the experiments presented in this section, we placed the splitting point after the third pooling layer of the VGG16 architecture.

\textbf{TD mechanisms.} In addition to NN-based approaches explained in Section \ref{subs:decision_network}, we consider the following alternatives:
\begin{itemize}
    \item We define the \textit{confidence} of the early exit classifier as the maximum value within the softmax output of the classifier. We transmit the features to the edge server only if the confidence is lower than a predefined threshold. We denote this approach as \textit{Confidence threshold}. This mechanism is closely related to the method proposed in \cite{split_early_exit}, which also utilizes confidence thresholds for making the decision.
    \item We first calculate the entropy of the softmax output of the classifier as:
    \begin{equation}
        H(x) = -F_e(x) \bullet \log_2(F_e(x)),
    \end{equation}
    where $\bullet$ indicates the dot product operation. We then define an entropy threshold and transmit only if the entropy for a given sample is lower than the threshold. A similar criterion has been utilized in \cite{branchynet_early_exit}. We denote this approach as \textit{Entropy threshold}.
    \item We first use the training split of the dataset to calculate the average expected accuracy and communication savings for each class under different confidence thresholds, as defined above, and multiple values of $\mathrm{SNR}$. During inference, we find the set of per-class thresholds that maximizes the objective criterion defined as a weighted sum of expected accuracy and communication savings. We denote this approach as \textit{Per-class confidence threshold}.
    \item \textit{Early exit}, which corresponds to a scheme, where the features are never transmitted to the edge server, and only the predictions made by the early exit are utilized. This corresponds to a case, where $D(x)=1$. The opposite of this scheme is \textit{Final exit}, which assumes the features are always transmitted to the edge server, and only the predictions made by the final exit are considered, which corresponds to $D(x)=0$.
    \item \textit{Random decision}, where the transmission of each sample is decided randomly using a uniform distribution.
\end{itemize}


\subsection{Performance comparison}
In this section, we compare the classification accuracy achieved by the TD methods described in Section \ref{subs:experimental_setup} and the TD NN trained with each of the strategies described in Section \ref{subs:decision_network}.
\begin{figure*}[ht]
    \centering
    \begin{subfigure}[t]{0.49\textwidth}
        \centering
        \includegraphics[width=\textwidth]{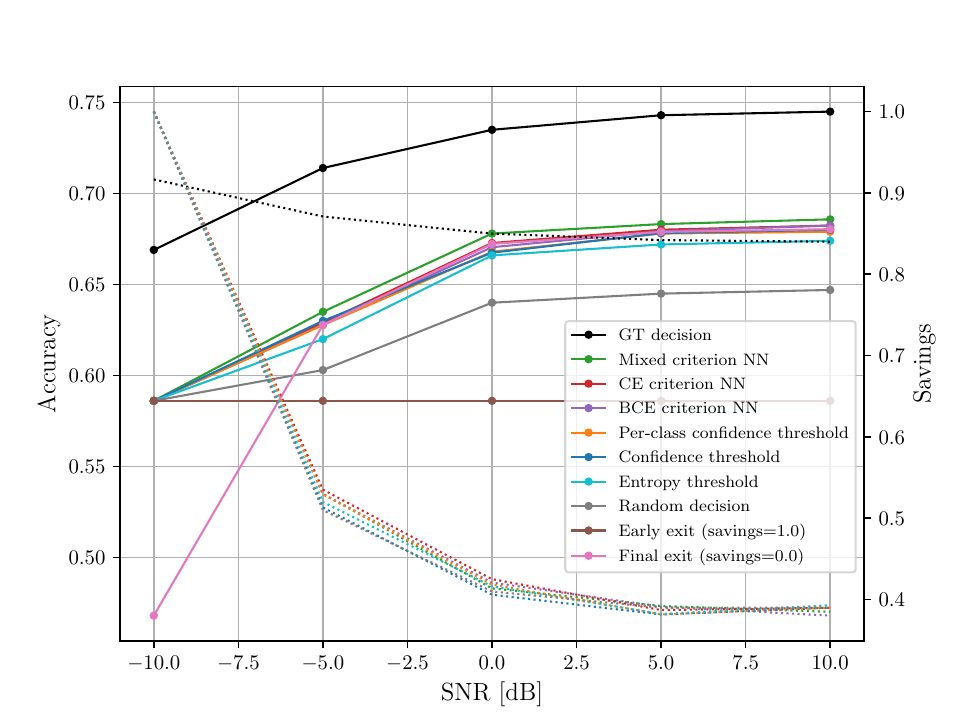}
        \caption{}
        \label{fig:acc_comparison_fixed_savings}
    \end{subfigure}
    \begin{subfigure}[t]{0.49\textwidth}
        \centering
        \includegraphics[width=\textwidth]{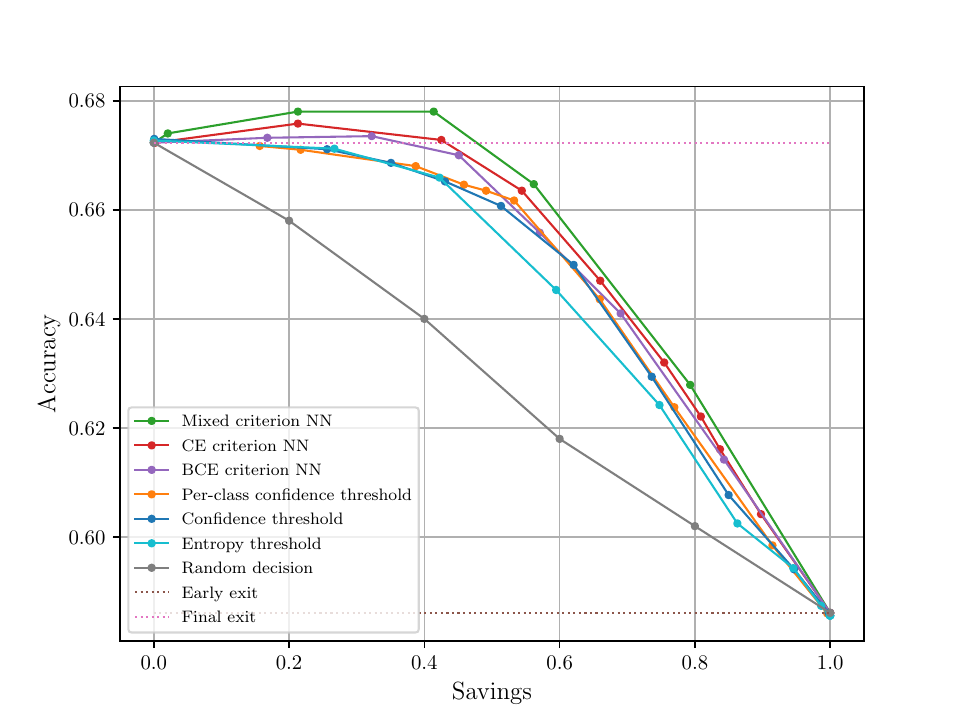}
        \caption{}
        \label{fig:acc_comparison_fixed_snr}
    \end{subfigure}
\caption{Comparison of the accuracy (solid lines) achieved by different models for (a) comparable communication savings (dotted lines) as a function of channel SNR, (b) fixed channel $\mathrm{SNR}=0\,\mathrm{dB}$ as a function of communication savings.}
\label{fig:acc_comparison}

\end{figure*}
In Fig. \ref{fig:acc_comparison_fixed_savings}, we plot the accuracy achieved by each of the methods as a function of the channel SNR. We tune the hyperparameters of each method to achieve similar communication savings for a given SNR. The \textit{Mixed criterion NN} approach, which is based on a TD NN trained with the loss function in Eq. (\ref{eq:mixed}), for $\alpha = 0.1$, slightly outperforms other approaches, most of which achieve similar results (which will be further analyzed in Section \ref{subs:ablation_study}), and the entropy threshold approach performs slightly worse than the confidence-based approaches. More importantly, significant communication savings (almost 45\% for $\mathrm{SNR} = 0\,\mathrm{dB}$) can be achieved by applying the majority of TD mechanisms studied in this work. We note that proper utilization of these mechanisms leads to achieving better or on-par performance compared to using only early exit or final exit, which validates the need for implementing dynamic early exiting mechanisms in the context of edge collaborative inference systems. Another observation is that the random decision, as expected, achieved significantly lower accuracy than the other approaches. The \textit{GT decision} approach, which refers to the ground truth, and only transmits when the early exit is incorrect and the final exit is correct, serves as an upper bound. It consistently allows more than $85\%$ savings at an accuracy significantly higher than the other approaches. One can observe that savings decrease as the SNR increases since less noise leads to better accuracy at the final exit, making it more beneficial to transmit the intermediate features to the edge server. The TD mechanisms studied in this work can take that information into consideration, and produce decisions that can accurately assess whether additional processing at the edge server is required.

In Fig. \ref{fig:acc_comparison_fixed_snr} we first fixed SNR to $0\,\mathrm{dB}$ and varied the hyperparameters of each of the methods to achieve different values of communication savings. Again, slightly improved accuracy was achieved by Mixed criterion NN, closely followed by \textit{CE criterion NN}, where the TD NN was trained with the loss function defined in Eq. (\ref{eq:joint}). The other methods achieved similar accuracy for comparable communication savings levels. Early exit and Final exit are visualized as dotted lines on the figure (Early exit savings are always equal to $1.0$). We can see that at savings of approximately $0.4$, the Mixed scheme is able to outperform Final exit by $0.6\%$ in accuracy, which proves that avoiding the transmission and utilizing a less powerful early exit can bring communication savings and provide better quality predictions. Another important observation is the fact that the Mixed approach NN is effectively a combination of the CE criterion NN and \textit{BCE criterion NN} approaches, yet it outperforms both of them. We hypothesize that the reason for this behavior is that the CE criterion NN effectively learns to put more weight on the output that is more likely to produce the correct prediction in given circumstances. A small addition of the binary cross-entropy term from the BCE criterion NN further avoids transmission when both early and final exits are not likely to produce a correct prediction. This leads to further communication savings without imposing any accuracy loss.

\begin{figure}[t]
    \centering
\includegraphics[width=0.7\textwidth]{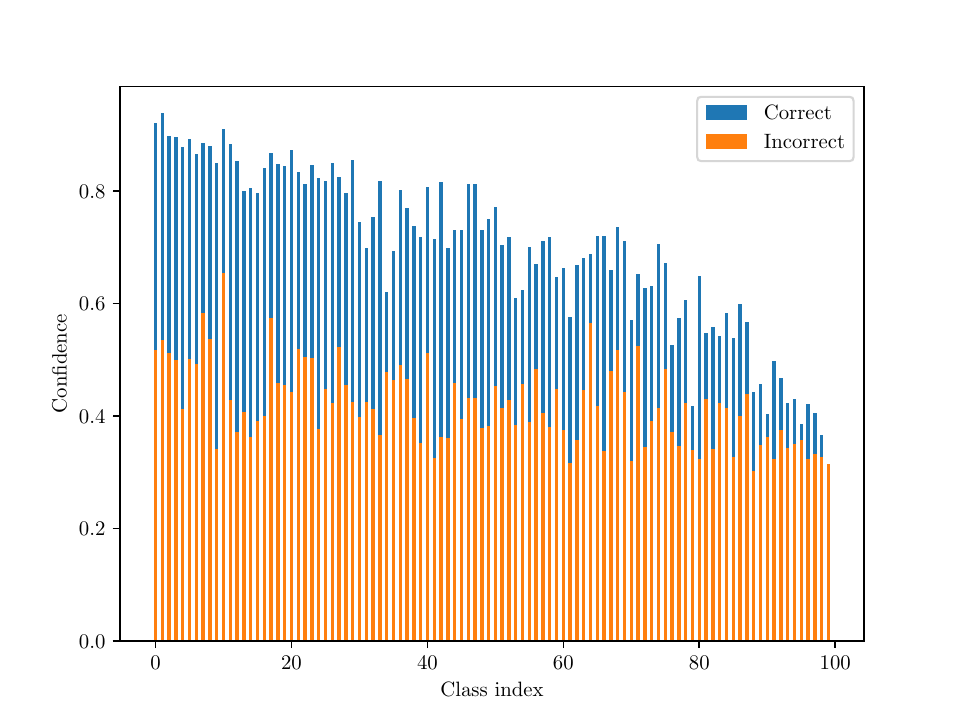}
    \caption{Average per-class confidences of the early exit predictions for the correct and incorrect predictions for all the classes in the CIFAR100 dataset. In general, the confidences are higher for the correct predictions; however, the gap may differ significantly between classes.}
    \label{fig:confidencies_early_true_false}
\end{figure}

\subsection{Ablation study}
\label{subs:ablation_study}
\textbf{Confidence analysis.} In this section we analyze the distribution of the per-class confidences of the early exit outputs separately for the correct and incorrect predictions (please see Fig. \ref{fig:confidencies_early_true_false}). We note that the confidences of the correct predictions are generally higher than the confidences of the false predictions, yet the exact ratio between the two may vary significantly from as high as $2:1$ to $1:1$ for different classes. This leads to the conclusion that information about the classes has to be taken into consideration when making the decision. Surprisingly, when analyzing Fig. \ref{fig:acc_comparison}, one can quickly notice, that the Per-class confidence threshold and Confidence threshold schemes exhibit significant overlap, despite the latter utilizing only a single threshold value for all the classes. The reason for that behavior will be analyzed next.

\begin{figure}[t]
    \centering
    \includegraphics[width=0.7\textwidth]{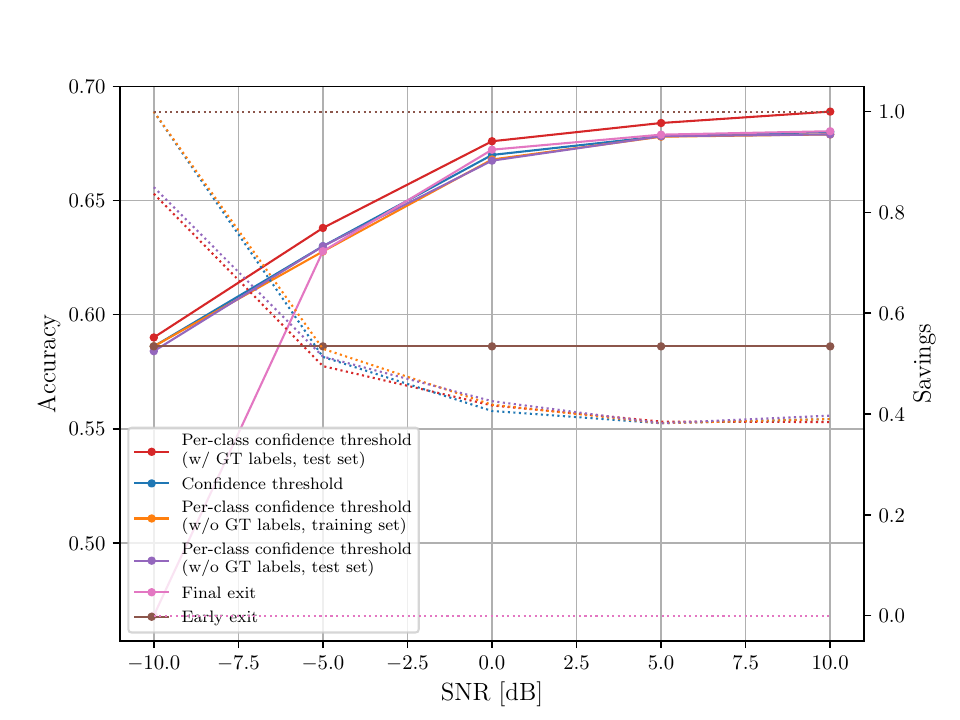}
    \caption{Comparison of the accuracy (solid line) and savings (dotted line) of the methods based on early exit predictions confidence. Hyperparameters of each method were tuned to provide comparable savings at each SNR.}
    \label{fig:acc_comparison_thresholds}
\end{figure}

\textbf{Performance of the confidence threshold methods.} In Fig. \ref{fig:acc_comparison_thresholds} we compare the accuracy achieved by various confidence-based threshold methods for different values of channel SNR and comparable savings. We note that the accuracy of the method that sets separate thresholds for each class, based on the training set performance, is similar to that achieved by the method that applies a single threshold to all the classes. To further investigate the reason for this behavior, we obtained a new set of per-class confidence thresholds. As opposed to the original method, the new thresholds were found based on the test set of the CIFAR100 dataset (indicated as \textit{test set} in the label). We further noted that the class with the highest softmax score assigned by the early exit is not necessarily the correct one. Therefore, for the sake of comparison, we consider another scheme, which selects the proper per-class threshold based on the ground truth label. This ensures the correct per-class threshold is selected each time during inference. We see that only in the case of utilizing the test set to calculate the thresholds and ground truth labels to select the correct threshold during inference, the per-class confidence method was able to achieve improvement over the single threshold method. This indicates that using the training set to find optimal thresholds, as well as the error incurred by selecting thresholds based on the most confident class may mislead the TD mechanism and cause a significant accuracy drop of the per-class confidence threshold method.

\begin{figure}[t]
    \centering
    \includegraphics[width=0.7\textwidth]{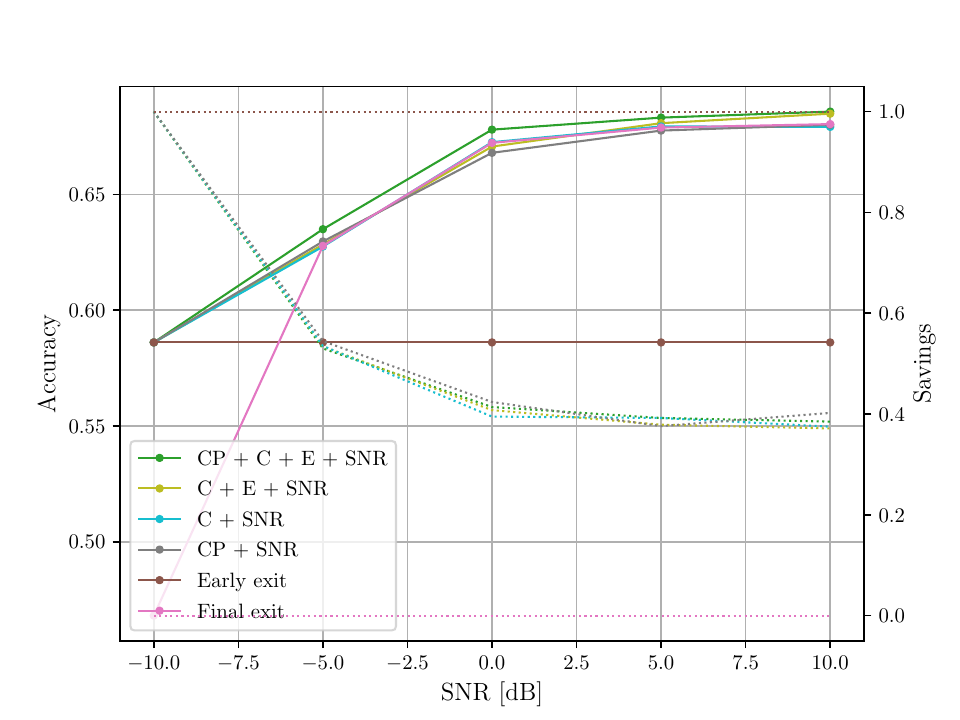}
    \caption{Comparison of the accuracy (solid line) and savings (dotted line) of the Mixed criterion NN, given different inputs to the TD NN, as a function of SNR. Hyperparameters of each method were tuned to provide comparable savings for each SNR. Early exit class probabilities (CP), confidences (C), entropy (E) of the class predictions, and SNR were used as inputs. The best-performing combination includes all of the inputs.}
    \label{fig:acc_comparison_fixed_savings_ablation}
\end{figure}

\textbf{Optimal TD NN inputs.} The above analysis (Fig. \ref{fig:confidencies_early_true_false}) about the necessity of considering the entire distribution of class predictions rather than some high-level statistical information about it has been experimentally validated. In Fig. \ref{fig:acc_comparison_fixed_savings_ablation} we show the comparison of the accuracy of the models, where the TD NN was employed and trained with the Mixed criterion. The architecture of the TD NN and other hyperparameters remain unchanged, and we only modify the inputs to the network to find their impact on the accuracy. We see from the results that providing only the channel SNR and the class probability (CP) distribution of the early exit is not sufficient to achieve satisfactory performance. The behavior is similar if we consider SNR and the statistics that summarize the CP distribution, including confidence (C) and entropy (E). The best result is achieved by combining all the inputs, i.e., CP distribution, C, E, and SNR. We note that the information about the entropy and the confidence is already implicitly included in the CP distribution, but we hypothesize, that the TD NN operates best when both high- and low-level information is provided to it as input.

\textbf{Complexity analysis.} In Table \ref{tab:complexity} we compare the complexity of the TD NN to the other parts of the DNN. We note that in the scenario studied in this work, i.e., for VGG16 DNN with a partitioning point selected after the third average pooling operation, the complexity of TD NN is small. When we compare the complexity of the part of the DNN deployed at the edge device with the complexity of the TD NN, we can see that the latter is responsible for approximately $0.01\%$ of the floating-point operations (FLOPs) performed at the edge device. We further argue, that the computational burden of 94000 FLOPs imposed by the TD NN, is acceptable given the modern hardware standards.

\begin{table}[t]
\caption{Complexity of the TD NN compared to the other parts of the DNN.}
\centering
\label{tab:complexity}
\begin{tabular}{|l|l|}
\hline
\textbf{Model}                       & \textbf{Complexity}     \\
 \hline \hline
TD NN          & 0.094 MFLOPs   \\ \hline
Early exit classifier head               & 0.025 MFLOPs   \\ \hline
DNN part at the edge device & 97.990 MFLOPs  \\ \hline
Full DNN                    & 320.717 MFLOPs \\ \hline
\end{tabular}
\end{table}

\section{Conclusions}
In this work, we presented an early exit mechanism for collaborative inference systems deployed at the wireless edge, where an image classification DNN is split into two parts and deployed between the edge device and the edge server. We introduced a TD NN, which utilizes the information provided by the early exit mechanism and the channel state to make a binary decision indicating whether the early exit output should be accepted, or the data should be offloaded to the edge server for further processing. We showed that, through careful design of the TD mechanism, either learnable or not, it is possible to achieve significant communication savings, while also outperforming strategies based on utilizing only early exit or final exit, yet the differences between the accuracy achieved by various TD mechanisms are not significant. Through further experimentation, we also illustrated that the entire early exit output distribution, combined with summarizing statistics should be used as an input to the TD mechanism in order to achieve satisfactory performance.

\bibliographystyle{IEEEtran}
\bibliography{refs}

%







\end{document}